\documentclass[sigconf,nonacm]{acmart}

\AtBeginDocument{%
  }

\setcopyright{acmlicensed}
\copyrightyear{2018}
\acmYear{2018}
\acmDOI{XXXXXXX.XXXXXXX}
\acmConference[Conference acronym 'XX]{Make sure to enter the correct
  conference title from your rights confirmation email}{June 03--05,
  2018}{Woodstock, NY}
\acmISBN{978-1-4503-XXXX-X/2018/06}

\usepackage{algorithm}
\usepackage[noend]{algpseudocode}
\usepackage{wrapfig}

\begin{document}

\title{Gandalf: Flash Shall Not Die! Increasing Flash Storage Lifetime for KV Cache Offloading in LLM Serving}

\settopmatter{authorsperrow=4}






\author{Maharani A. P. Irawan}
\authornote{Part of this work was done while the author was an intern at TensorMesh Inc.}
\affiliation{%
  \institution{The University of Chicago}
  \country{USA}
}

\author{Kuntai Du}
\affiliation{%
  \institution{TensorMesh Inc.}
  \country{USA}
}

\author{Jiayi Yao}
\affiliation{%
  \institution{The University of Chicago}
  \country{USA}
}

\author{Dongjoo Seo}
\affiliation{%
  \institution{Samsung Semiconductor}
  \country{USA}
}

\author{Yuyang Huang}
\affiliation{%
  \institution{The University of Chicago}
  \country{USA}
}

\author{Haryadi S. Gunawi}
\affiliation{%
  \institution{The University of Chicago}
  \country{USA}
}

\author{Junchen Jiang}
\affiliation{%
  \institution{TensorMesh Inc.}
  \country{USA}
}
\additionalaffiliation{%
  \institution{The University of Chicago}
  \country{USA}
}

\begin{abstract}

KV caching reduces the serving cost of large language models (LLMs), but the growing KV cache footprint is increasingly exceeds GPU and host-memory capacity. 
Therefore, people use SSD to expand the KV cache storage capacity with high I/O bandwidth.
However, in production LLM workload, such as multi-round conversation and LLM agents, the total amount of KV cache writes significantly exceeds SSD capacity, shortening SSD lifetime from years to months.
To improve SSD lifetime, a natural solution is to reduce the KV cache writes by gating the KV cache writes via historical access patterns.
However, that significantly lowers the KV cache hit rate (and increase the LLM serving cost) since KV caches from different attention types might have identical historical access pattern but different future access patterns.
Our insight is that, to reduce KV cache writes without hurting KV cache hit rate, one need to not only consider the access pattern, but also application-level information such as the attention type of KV cache.
Leveraging this observation, we propose \sys, the first SSD KV cache offloading system with SSD life time awareness.
\sys prioritize the admission of full-attention KV cache into SSD over the cache generated by other types of attention, which significantly increases SSD lifetime without sacrificing cache hit rate.
Our evaluation shows that \sys extends the SSD lifetime by 7.3\tms on average, while having negligible impact on cache hit rate across multiple workload traces and wide range of hardware capacity.
\sys design can be extended to other flash-based storage, such as High Bandwidth Flash (HBF).

\end{abstract}

\def \sys {\textsc{Gandalf}\,}

\def \ie {\textit{i.e.}~}
\def \eg {\textit{e.g.}~}
\def \etc {\textit{etc.}~}
\def \tms {$\times$~}
\def \dv {$\div$~}

\newcommand{\cmark}{\ding{51}}
\newcommand{\xmark}{\ding{55}}

\definecolor{Creamy}{RGB}{255, 105, 180}
\newcounter{ranicounter}
\setcounter{ranicounter}{1}
\newcommand{\rani}[1]{{\footnotesize
\textbf{\textcolor{Creamy}{(RANI$_{\arabic{ranicounter}}$: #1)}}}
\stepcounter{ranicounter}}

\definecolor{ForestGreen}{RGB}{34,139,34}
\newcounter{kuntaicounter}
\setcounter{kuntaicounter}{1}
\newcommand{\kt}[1]{{\footnotesize
\textbf{\textcolor{ForestGreen}{(KUNTAI$_{\arabic{kuntaicounter}}$: #1)}}}
\stepcounter{kuntaicounter}}

\definecolor{CornflowerBlue}{RGB}{100,149,237}
\newcounter{jiayicounter}
\setcounter{jiayicounter}{1}
\newcommand{\jiayi}[1]{{\footnotesize
\textbf{\textcolor{CornflowerBlue}{(JIAYI$_{\arabic{jiayicounter}}$: #1)}}}
\stepcounter{jiayicounter}}

\definecolor{Brown}{RGB}{150,75,0}
\newcounter{jccounter}
\setcounter{jccounter}{1}
\newcommand{\jc}[1]{{\footnotesize
\textbf{\textcolor{Brown}{(JC$_{\arabic{jccounter}}$: #1)}}}
\stepcounter{jccounter}}

\maketitle

\section{Introduction}
\label{sec-intro}



Large language models (LLMs) have demonstrated strong capabilities in multi-round conversations and agentic workflows, garnering wide adoption.
These workloads accumulate conversation histories, tool outputs, and intermediate states, resulting in increasingly long and frequently reused contexts.
However, recomputing the attention key-value (KV) states for these contexts at every rounds substantially increases inference latency and cost.
Hence, people store all KV caches in their system to avoid recomputation and increase cache hit rate~\cite{web.openaiPromptCaching, web.claudePromptCaching}.
%
%
%
%
%
Production workloads generates Terabytes of KV-cache per day at the amount that exceeds the capacity of GPU and host memory, motivating the use of higher-capacity storage tiers.
Flash-based storage, \eg NVMe SSDs, provides a scalable and comparatively cost-effective tier for storing these caches~\cite{web.nvidiaHierarchy}.


\begin{figure}[t]
    \centering
    \includegraphics[width=170pt]{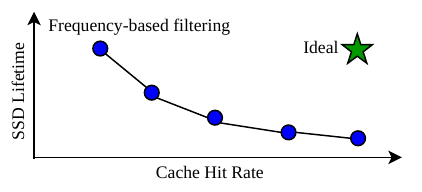}
    \caption{Using historical access pattern creates a trade-off between cache hit rate and SSD lifetime. \sys maintains cache hit rate while increasing SSD lifetime.}
    \label{fig-intro-tradeoff}
\end{figure}

This additional capacity comes with an endurance challenge because SSDs can only handle a finite volume of writes before they are worn out~\cite{brewer2011nonvolatile}.
Under production workloads, aggressive KV-cache admission improves cache-hit rate but generates substantial write volume, exhausting SSD endurance within days to months.
Therefore, existing production-level systems came up with simple filtering techniques that gates SSD admission, trading potential cache hits for fewer writes.
Frequency-based solutions, for example, offloads a KV cache to SSD only if it has demonstrated $N$\tms reuse, with $N$ as a hyperparameter~\cite{web.nvidiaKVBM, peng2026lruk}.
However, even with such filtering, the aggregate KV-cache write volume remains excessive for practical SSD endurance.
%
%
%
%
This raises a central question: 

\begin{center}
\textit{How can we improve the SSD lifetime without sacrificing KV caching performance benefit?}
\end{center}




\begin{table}[t]
\centering
\small
\caption{Reuse pattern of first round's KV caches in multi-round conversation}
\label{tab-mot-reuse-pattern}
\begin{tabular}{lcccc}
\toprule
Type & 2nd-round & 3rd-round & Future rounds & Reuse count \\
\midrule
Full attn & \cmark & \cmark & \cmark & $N_{\mathrm{rounds}}$ \\
Linear attn & \cmark & \xmark & \xmark & 1 \\
\bottomrule
\end{tabular}
\end{table}

The key challenge lies in predicting the future reuse of KV caches such that we can selectively retain only the most reusable caches to make the most of SSD's capacity.
Future prediction is difficult, but we found application-specific property that strongly correlates with reuse pattern.
We leverage two observations: 
(1) Recent LLMs increasingly employ hybrid architectures that combine full-attention layers with linear- or sparse-attention layers, and
(2) These different attention layer types produce KV caches with substantially different reuse patterns under emerging mainstream workloads: multi-round conversations and agentic workflows.
In particular, states from linear- and sparse-attention layers are reused less frequently than KV caches from full-attention layers. 
%
%
%
%
%


Consider a multi-round conversation workload use-case, which reuse pattern illustrated in {\bf Table~\ref{tab-mot-reuse-pattern}}.
At the beginning of each new round, the full-attention KV entries for the entire preceding context is needed. This pattern repeats for every new round, hence these KV entries are reused across all future rounds.
In contrast, a linear-attention layer's state and sliding-window cache is only needed to generate the immediate next round's response, and is then discarded, hence every state is only reused once.


This reuse pattern heterogeneity motivates attention-type-aware cache placement.
Writing large, one-time use linear-attention states to SSD provides little cache hit benefit while consuming both write endurance and cache capacity.
Layer type can therefore serve as a practical placement signal, allowing us to utilize SSDs to expand cache capacity while minimizing endurance loss.
%
%
We propose an attention layer-aware placement policy that caches short-lived linear-attention states in host memory while keeping longer-lived full-attention KV caches on SSD.


However, there are 2 practical challenges that entail this placement policy.

\textit{First}, this placement policy is too rigid: the size of CPU DRAM and SSD is fixed, but workload can generate different amounts of different attention states.
Particularly, a workload with many concurrent sessions can generate a large number of reusable linear-attention states that exceeds the available DRAM capacity, similarly, a workload with long contexts can generate a large number of reusable full-attention KV caches that exceeds the available SSD capacity.
We call this a hardware-to-utilization imbalance.
This imbalance forces frequent cache eviction on the saturated tier despite available capacity elsewhere in the storage hierarchy, thus lowering cache-hit rate.
Worse, if the overutilization happens in SSD, the resulting write churn can consume its endurance budget more rapidly and further shorten its lifetime.
Therefore, we design a capacity-aware placement policy that filters SSD admission based on utilization and pressure of each tier, balancing cache hit rate against SSD endurance.



\textit{Second}, such rigid placement moves cache hit from DRAM to SSD, which can hurt TTFT because SSD bandwidth is substantially lower than DRAM bandwidth.
Hence, transferring large full-attention KV caches from SSD may hurt TTFT.


We explore the potential of attention-type-aware placement using an offline oracle that jointly optimizes SSD admission and cache eviction decisions, biased towards our tier placement.
The oracle reveals two complementary design principles.
\textit{First}, placement must adapt to tier capacity and selectively persist KV caches on SSD to preserve the prefix-hit boundary.
\textit{Second}, admission and eviction must be optimized jointly: admitting a new cache has cost and benefit. 
It benefits future reuse, but it has write cost and replacement cost---cost of replacing other cache entries, may or may not already pay for its promised benefit.
An SSD admission is beneficial only when its expected reuse value exceeds the sum of write cost and the value of the entries it evicts.
Together, these principles characterize the best attainable trade-off among SSD lifetime, cache-hit rate, and TTFT.
We then distill the oracle's decisions into an online, capacity-aware policy that estimates future reuse from historical access patterns.






\textbf{Contributions.} Our work makes the following contributions:
\begin{enumerate}

\item We are the first to show that the SSD lifetime is significantly lower than its factory guarantee in production-scale SSD-based KV cache offloading.


\item We show that, leveraging application-level information (the attention type corresponds to each KV cache objects), we can significantly reduce the write volume and increase SSD life time with negligible impact on KV cache hit rate.


\item We propose \sys, a KV caching system that can increase SSD lifetime by 7.3\tms (on average) with negligible impact on hit rate across a wide range of  LLMs and hardware capacity.



\end{enumerate}


Although this paper only demonstrates \sys on SSD, but the technique is extendable to other flash-based storage devices, such as High Bandwidth Flash (HBF).

\section{Motivation}
\label{sec-background}

\subsection{KV Caches Exceed Memory Capacity}



Over the past years, the increasing intelligence and context window size of LLMs empowered multi-round conversations and agentic workflows.
These types of workloads continuously accumulate conversation history, intermediate reasoning, tool outputs, and subagent results.
Unlike single-turn requests, these workloads repeatedly extend and reuse an existing prefix, causing each session's KV-cache footprint to grow throughout its lifetime.
Hence, the aggregate KV cache sizes saved across requests and sessions are getting enormous.


Long contexts also make cache misses increasingly expensive.
When a reusable prefix is not cached, the serving system must recompute its KV states by prefilling all preceding tokens before generating a response.
In workloads with long context sizes, such as multi-round conversation, agentic workflow, and long document processing, this repeatedly consumes scarce GPU computation, increasing TTFT and serving cost.
People therefore seek to retain as much reusable KV state as possible, effectively trading storage capacity for computation.



\begin{table}[t]
\centering
\small
\caption{
Under KV cache offloading workloads, SSDs may last less than a year—far short of their five-year warranty.
}
\label{tab-mot-ssd}
\begin{tabular}{llrr}
\toprule
Workload & Model & \begin{tabular}[b]{@{}r@{}}KV cache\\size (TB)\end{tabular} & SSD lifetime (y) \\
\midrule
\begin{tabular}[t]{@{}l@{}}Claude Code\\\texttt{cc-traces-062126}\end{tabular}
 & Qwen3.5-27B & 98 & 0.4 \\
 & Kimi-K3 & 240 & 0.2 \\
\begin{tabular}[t]{@{}l@{}}Claude Code\\\texttt{cc-traces-060826}\end{tabular}
 & Qwen3.5-27B & 81 & 0.4 \\
 & Kimi-K3 & 199 & 0.2 \\
\bottomrule
\end{tabular}
\end{table}

Our analysis of two one-day SemiAnalysis Claude Code traces~\cite{semianalysisai_cc_traces_weka_062126} estimates that representative hybrid models generate 81--240~TB of KV-cache state per day ({\bf Table~\ref{tab-mot-ssd}}). 
At peak, the concurrently active KV-cache working set reaches 8.6TB, exceeding practical host-memory capacity, hence provisioning sufficient CPU DRAM would be prohibitively expensive.
Retaining this state for future reuse is therefore infeasible using memory alone. 
Consequently, the cache capacity must extend beyond GPU and host memory to higher-capacity local storage.



\subsection{Why not HDD cluster?}
\label{sec-mot-hdd}

Local NVMe SSD is a popular extension of the memory hierarchy for LLM serving~\cite{rasley2020deepspeed,sheng2023flexgen, gao2024cachedattention, jeong2025flashgen, noohi2026tenure, tumkur2026whereshouldkvcachelive}.
Since SSD is known to have a limited endurance problem, why haven't people used HDD instead?
To make HDD fast enough, we could parallelize I/O across a cluster of HDDs.
To answer this question, we first compare the endurance of SSD and HDD given the manufacturer's specifications.


{\bf SSD} stores data by programming charge in NAND-flash cells, which tolerate only a finite number of program/erase (write) cycles.
Manufacturers quantify this endurance as Terabytes Written (TBW), or equivalently Drive Writes Per Day (DWPD) over a warranty period.
For a drive with capacity $C$, rated for $D$ DWPD over $Y$ years, its host-write endurance budget is $E_\mathrm{TBW} = C \times D \times 365 \times Y$.
%
Garbage collection can amplify host writes by copying live data while reclaiming space.
The write-amplification factor (WAF) measures this additional write traffic: $\mathrm{WAF} = \frac{W_{\mathrm{NAND}}}{W_{\mathrm{host}}}$, where $W_{\mathrm{host}}$ and $W_{\mathrm{NAND}}$ denote the logical host writes and physical NAND writes, respectively.
Thus, SSD will live until
\begin{equation}
    T_{\mathrm{SSD}} = \frac{E_\mathrm{TBW}}{\mathrm{WAF}\times E_\mathrm{workload}} \text{ Terabytes Written}
\end{equation}

{\bf HDDs} store data by magnetizing regions of rotating platters.
Manufacturers specify an annual total read and write limit of $RW_{\mathrm{limit}}$~TB/year.
Over $Y$ years, this rate corresponds to a nominal cumulative transfer envelope of $E_{\mathrm{HDD}} = RW_{\mathrm{limit}} \times Y$.
Manufacturers also specify HDD reliability by Annualized Failure Rate (AFR) and Mean Time Between Failures (MTBF).
Exceeding $E_{\mathrm{HDD}}$ results in increasing AFR and decreasing MTBF, hence more frequent failures and recovery overhead will occur, slowing the HDD for both reads and writes.
For a KV-cache workload that writes $W_{\mathrm{host}}$ bytes and reads each written byte an average of $h$ times, the total HDD workload is $E_{\mathrm{workload}} = (1+h)W_{\mathrm{host}}$.
HDD operates reliably until
\begin{equation}
    T_{\mathrm{HDD}} = \frac{E_\mathrm{HDD}}{E_\mathrm{workload}} \text{ Terabytes Read and Written}
\end{equation}
This differs from SSDs, for which reads do not consume the rated TBW endurance.









\begin{figure}[t]
    \centering
    \includegraphics[width=170pt]{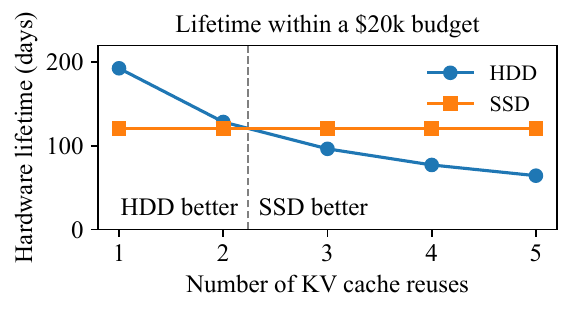}
    \caption{Under same acquisition budget, SSD is more optimal to be used for KV cache offloading workload, which is reuse-heavy.}
    \label{fig-mot-hdd}
\end{figure}

{\bf Takeaway:} Although HDDs avoid flash wear-out, both reads and writes exercise their mechanical components and count toward their lifetime.
KV-cache workloads generate substantial traffic in both directions, making HDDs poorly suited as a storage.
{\bf Figure~\ref{fig-mot-hdd}} compares SSD and HDD configurations with equal aggregate I/O bandwidth and \$20k Total Cost of Ownership (TCO) budget. 
HDDs require substantially more devices, power, and space, yet it has a shorter projected lifetime than SSD when each written KV-cache entry is reused at least $3\times$.

Since multi-round conversation and agentic workloads trigger heavy reuse as shown in {\bf Table~\ref{tab-mot-reuse-pattern}}, SSD is the more cost-effective option compared to HDD, at the same Total Cost of Ownership (TCO).
Therefore, we focused on SSD for this work, and leave the integration with HDD for future work.
Although we focus on commodity enterprise SSDs, our placement and endurance-management techniques also apply to emerging flash-based context-memory tiers, including high-bandwidth flash (HBF).



\subsection{KV Caching is Rapidly Killing SSDs}








Under frontier model and agentic workloads, our analysis shows that even a 30-TB enterprise SSD can exhaust its rated TBW in less than one year.
Smaller enterprise SSDs wear out even faster, with a 2-TB drive reaching its rated TBW in just days.
%
%
In \S\ref{sec-mot-hdd}, we discussed that SSD wear is a product of host-write volume and device-level write amplification.
We target host writes by reducing the KV-cache writes issued to SSD.



Tiered KV caches commonly use write-through or write-back offloading.

\textit{Write-through} immediately copies every admitted KV states to SSD, maximizing availability at multiple tiers, but generating substantial SSD write.
This duplication preserves the entry after DRAM eviction, but wastes SSD endurance when all of its reuse occurs in a short time while it remains in DRAM.
The SSD copy then consumes write budget without serving a cache hit.

\textit{Write-back} persists demoted KV states from higher storage tier regardless of its expected reuse.
This policy still writes every DRAM eviction to SSD, including states that contribute little to future cache hits.

\textit{Selective write-through} reduces this traffic using a fixed reuse threshold; for example, NVIDIA Dynamo offloads a block only after its access frequency reaches two~\cite{web.nvidiaKVBM}.
However, frequency-only filtering ignores state size, layer-specific reuse, tier occupancy, and the remaining write budget, allowing large, short-lived states to consume substantial endurance for little benefit.


\subsection{How does prior work address SSD endurance?}

SSD endurance has long been a concern in flash-based caching systems.
Traditional systems, such as Flashcache~\cite{web.flashcache} and Google's Colossus Flash Cache~\cite{yang2022cachesack}, use SSDs to accelerate access to persistent data.
Recent studies show that KV-cache offloading introduces similar endurance challenges in LLM serving~\cite{son2026exploring,li2026hbfsucks}.
Existing approaches can be grouped into three categories.

\emph{Increasing physical write endurance.}
Hardware provisioning distributes writes across more SSDs, aggregating write budget and reducing the per-device write rate, thereby increasing aggregate write endurance at the cost of additional acquisition, power, and maintenance costs~\cite{lee2026prompt}.
Classic SSD endurance technique called retention relaxation increases tolerable NAND program/erase cycles by reducing data retention requirements~\cite{cai2012flash,luo2015warm}.
FlashAccel exploits the short lifetime of KV caches to trade retention for endurance~\cite{wang2026flashaccel}.
However, both approaches increase the available write budget rather than reducing host-level KV-cache writes, and may require additional hardware or device support.
They also do not leverage unique property of KV caches: we can choose to not cache the KV cache and recompute it when needed.
It may therefore remain insufficient for sustained LLM serving, especially in multi-round and agentic workloads.

\emph{Selective KV cache admission and eviction.}
Although not aimed at prolonging SSD lifetime, many existing cache admission and eviction policies selectively filters KV caches to optimize cache utilization and reuse to increase cache hit rate, indirectly reducing writes.
One approach schedules requests that involve resident KV caches, which KV caches fits in the available space~\cite{yu2026characterizinghighbandwidthflash}.
LRU-$K$ admits a KV cache only after it has been accessed $K$ times~\cite{peng2026lruk, lruk1993}.
DeepSeek-V4.1 retains sliding window KV caches in distributed memory pool to guarantee other KV cache availability for at least 72 hours inside the persistent storage~\cite{deepseekai2026deepseekv41flashpushinglimitskv}.
HotPrefix~\cite{li2025hotprefix} and PrefixShield~\cite{wang2026prefixshield} leverages prefix-based policies to evict cold KV caches from GPU HBM, while Workload-Aware~\cite{wang2025workloadaware} considers reuse probability, block position, and lifetime.
LPC~\cite{yang2026lpc} and SAECache~\cite{fang2026saecache} predict reuse through learning-based methods.
RR-evict~\cite{pan2026rr} uses a round-robin eviction to ensure fairness across sessions, while Bidaw~\cite{hu2026bidaw} predicts reuse distance from prior responses in the session.
Another approach uses recency, frequency, and EWMA-based reuse estimates for prefetching and eviction~\cite{tumkur2026whereshouldkvcachelive}.
MeritKV~\cite{khemani2026meritkv} gates admission to storage backend if the saved prefill time, minus lookup and transfer costs, minus historical miss' recomputation costs, is less than a threshold.
However, these approaches do not explicitly optimize SSD endurance.

\subsection{SSD lifetime and Cache Hit Rate Trade-off}



\begin{figure}[t]
    \centering
    \includegraphics[width=250pt]{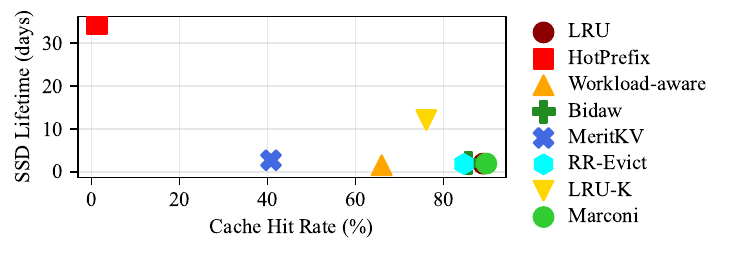}
    \caption{Existing systems trades off SSD lifetime and cache hit rate.}
    \label{fig-eval-tradeoff}
\end{figure}

Hardware configuration---size of host memory and SSD capacity---is fixed.
But workload demand can vary substantially over time.
Since cache entries are reused homogeneously, the storage hierarchy must treat them uniformly.
Hence, existing solutions trades off SSD lifetime against cache hit rate, as shown in {\bf Figure~\ref{fig-eval-tradeoff}}.


Storing more KV-cache entries on SSD increases the probability that future requests find reusable state, thereby improving the cache-hit rate and reducing TTFT.
However, the resulting write volume rapidly consumes the SSD's finite endurance budget.
%
%
Conversely, storing fewer entries reduces SSD traffic and extends device lifetime, but may discard reusable state, lowering the cache-hit rate and increasing TTFT.
Thus, when cache entries are homogeneous, improving one objective necessarily comes at the expense of the other.
Breaking this trade-off requires identifying writes that consume SSD endurance but contribute little to future cache reuse.

\subsection{Key Observation: Hybrid model has heterogeneous reuse pattern}
\label{subsec-insight}


Dense models use full attention across all layers.
When a cached context is resumed, the KV entries of every tokens in the sequence prefix are reused.
%
%
Thus, storing a larger portion of the prefix allows more recomputation to be avoided, creating a near one-for-one trade-off between SSD writes and cache-hit rate.
Each layer and token's admitted KV entry provides comparable reuse value, and reducing SSD writes generally sacrifices a corresponding caching opportunity.
This creates the apparent one-for-one trade-off between SSD lifetime and cache-hit rate.


Hybrid models break this trade-off due to its model architecture: combining full-attention layers with sparse attention layers, such as linear and sliding-window attention.
As shown in {\bf Table~\ref{tab-mot-reuse-pattern}}, cached states from different layer types exhibit substantially non-uniform reuse patterns, where linear state is reused much less than full attention KV cache.
We observed that under append-only workloads such as multi-round conversation, all prefix tokens' full-attention KV caches are reused in every subsequent round, so their reuse count increases as the conversation grows.
By contrast, each sparse-attention state is typically used only once---to generate the immediately following round---before being superseded by a newer state. 


Uniformly writing these short-lived states to SSD is inefficient.
Write-through immediately persists them, while write-back may persist them upon eviction, even though they are likely to be superseded after only one reuse.
When the latest linear-attention state can remain in DRAM until the next round, its SSD copy provides no additional cache hit but still consumes SSD endurance and capacity.
Such writes may also evict more reusable full-attention KV entries, further reducing cache effectiveness.





\section{Overview}

\begin{figure}[t]
    \centering
    \includegraphics[width=230pt]{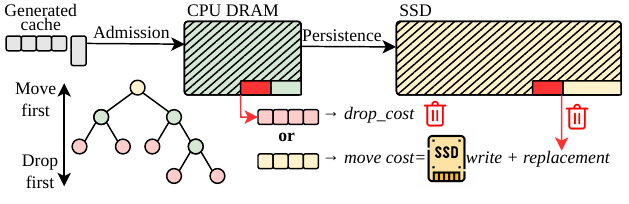}
    \caption{\sys overview. \sys admits new KV caches to host memory and, when it is full, chooses between dropping a DRAM cache or moving it to SSD based on SSD write and replacement-cache recomputation costs.}
    \label{fig-ovv-terms}
\end{figure}

As shown in {\bf Figure~\ref{fig-ovv-terms}}, \sys involves a two-tier storage hierarchy setup, comprising of DRAM (L1) and SSD (L2).
{\bf \S~\ref{subsec-placement}} describes our attention type-aware placement policy that strictly decides the placement of each admitted state based on the attention type's reuse pattern.
{\bf \S~\ref{subsec-understanding}} investigates the potential of attention-type-aware placement by an oracle-based optimal algorithmic approximation that maximizes cache hit rate while minimizing SSD writes in a wear-aware manner.
As a result of the investigation, {\bf \S~\ref{subsec-capacity}} explains the capacity-aware cache management policy that decides whether to drop or move the state to SSD based on its expected future reuse and the SSD write cost consumed.
{\bf \S~\ref{subsec-optimization}} explains an optimization to reduce the score computation overhead while maintaining its effectiveness.

\subsection{Attention type-aware placement}
\label{subsec-placement}



Hybrid models produce cached states with substantially different sizes and reuse patterns.
Full-attention layers retain per-token KV entries, so their cache grows with the prefix length, whereas linear-attention layers maintain a fixed-size recurrent state at each checkpoint.
Sliding-window layers retain only the most recent $W$ tokens and can therefore use the same placement policy as linear-attention layers.
The state types also differ substantially in size. Across our evaluated models, a single linear-attention state occupies as much space as the full-attention KV entries for $2k$--$16k$ tokens.
However, reuse is only meaningful when all types of states are available at the same prefix boundary.



This heterogeneity in size and reuse pattern brings 2 opportunities: (1) not all KV state needs to be cached, and (2) where an admitted state should be placed.
The heterogeneity creates two distinct cache-management decisions.
First, \emph{admission} determines which generated states provide sufficient reuse value to retain.
Second, \emph{placement} determines the storage tier in which each admitted state should reside.


Prior works has solved \textit{``which KV state should be cached''} by probing the reuse boundary~\cite{pan2025marconi, wang2026prefixshield}.
These mechanisms address which states should be retained, but they do not determine whether an admitted state should reside in DRAM or in SSD.
Our work is aimed to improve SSD lifetime by managing KV cache placement leveraging its attention layer type to give a hint on the most suitable placement in the storage tier.

We therefore retain linear state in DRAM and full attention KV caches in SSD.
Each linear state generated at the end of each conversation round is put in DRAM while the full attention KV cache is put on SSD.
When the storage tier (DRAM/SSD) is out of space, the least-recently-used cache on the tier is evicted.

\subsection{Understanding the Potential}
\label{subsec-understanding}


{\bf Mathematical model} to estimate the SSD-lifetime improvement of attention-type-aware placement over conventional write-back, independent from eviction strategies.
Consider a workload containing $s$ sessions, each with an average of $n$ cached tokens and $r$ conversation rounds.
Let $f$ denote the size of full-attention KV entries per token, and let $\ell$ denote the size of one linear-attention state.
The size of full-attention and linear-attention state for each session are therefore $F = nf$ and $L = r\ell$, respectively.

Under write-back, both full-attention KV entries and linear-attention states are admitted to DRAM and written to SSD upon eviction.
Let $C$ denote the CPU DRAM capacity reserved for KV cache.
Assuming each evicted byte is written to SSD at most once except the last $C$ bytes, the SSD host-write volume is $W_{\mathrm{write-back}} = s(F+L)-C$.
%
%

With attention type-aware placement, only a fraction $p\in[0,1]$ of the linear-attention states needs to be spilled to SSD, while all full-attention KV entries are written to SSD.
Hence, the SSD host-write volume is $W_{\mathrm{type-aware}} = sF + psL = s(F+pL)$.

SSD's projected lifetime $T$ is inversely proportional to its write rate since writing to SSD reduces its lifespan.
So the approximate SSD-lifetime improvement is
%
\begin{equation}
    I_{\mathrm{type-aware}}
    = \frac{T_{\mathrm{type-aware}}}{T_{\mathrm{write-back}}}
    \approx \frac{W_{\mathrm{write-back}}}{W_{\mathrm{type-aware}}}
    = \frac{s(F+L)-C}{s(F+pL)}
    \label{eq:lifetime-improvement}
\end{equation}
Thus, the attainable SSD-lifetime improvement grows with the size of linear-attention state relative to full-attention KV cache and decreases with the fraction of linear state that must spill to SSD.


{\bf Capacity requirement for type-aware placement.}
The ideal type-aware placement keeps linear-attention states in DRAM and full-attention KV cache in SSD. 
In Equation~\ref{eq:lifetime-improvement}, this corresponds to $p=0$, which maximizes the SSD lifetime improvement.
However, achieving this upper bound requires sufficient DRAM capacity to retain the live linear states.


\begin{figure}[t]
    \centering
    \includegraphics[width=200pt]{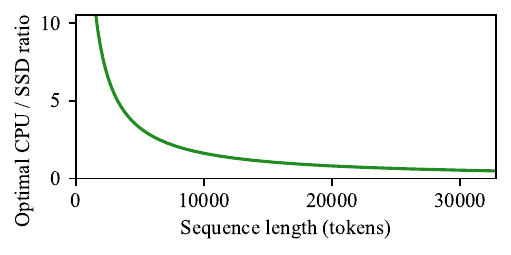}
    \caption{CPU/SSD capacity ratio to obtain maximum SSD lifetime improvement.}
    \label{fig-ovv-improv}
\end{figure}

As observed in Section~\ref{subsec-insight}, in multi-round conversation and agentic workloads, each session only needs to retain one linear-attention state at a time, but it must retain all full-attention KV entries for the entire prefix.
Strict separation requires enough DRAM to hold one live linear-attention state for every session and enough SSD capacity to hold every session's full-attention KV cache.
%
%
As shown in Figure~\ref{fig-ovv-improv}, with the same number of sessions, the CPU/SSD capacity ratio required to achieve maximum SSD-lifetime improvement decreases with session length because the live linear state has fixed size, whereas full-attention KV grows with the number of tokens.
Consequently, strict type-based placement cannot always use both tiers efficiently.
%
%
For short sessions, linear states dominate capacity and some may need to borrow SSD space.
%
%
For long sessions, full-attention KV dominates and may benefit from borrowing unused DRAM space.
%
%
We therefore design a capacity-aware placement algorithm as an SSD admission gate that treats attention type as a reuse pattern proxy for placement preference.
The gate selects which objects to move, retain, or evict based on their size, attention type-specific expected reuse, tier occupancy, and SSD-write cost.
We first study an offline optimal algorithm, approximate its decisions, and then derive a practical online policy.


{\bf Oracle-based approximation.}
We know that we can improve the SSD lifetime by $I_{\mathrm{type-aware}}$, but this model may not preserve the cache hit rate.
We therefore modeled an offline oracle condition that optimizes cache hit rates while minimizing SSD write volume to prolong SSD lifetime.
Sparse states are still kept in DRAM, while full KV caches are kept in SSD.

\textbf{Cost model.}
We keep two costs of admission: the benefit and the cost.
\textit{Benefit:} Promoting an object that is hit $H$ times saves $C_r \cdot H$ dollars of recomputation, where $C_r$ is the portion of GPU rent to recompute one chunk of KV.
\textit{Cost:} Promoting an object of size $s$ bytes costs $C_w \cdot s$ dollars, where $C_w$ is the portion of SSD purchase cost to write one byte.
Since SSD is eventually broken due to the write volume, we need to consider the cost of writing.
When an admission causes eviction of another object that still has remaining $H_{\mathrm{left}}$ hits, the benefit paid by the evicted object becomes the replacement cost of the admitted object.
Hence, the total cost becomes $C_w \cdot s + C_r \cdot H_{\mathrm{left}}$.

We want to maximize the net benefit, which is equivalent to maximizing the number of hits minus the consumed write volume:
\begin{equation}
    \max\; C_r\,\bigl(H - H_{\mathrm{left}}\bigr) - C_w\,W
    \;\Leftrightarrow\;
    \max\; \bigl(H - H_{\mathrm{left}}\bigr) - \lambda\,W,
    \;\;
    \lambda = \frac{C_w}{C_r},
    \label{eq:oracle-cost}
\end{equation}
where $H$ and $H_{\mathrm{left}}$ are summed over all promotions and $W$ is the total bytes written to SSD.
$\lambda$ tells us how many hits a written byte must earn to be worth writing.

\textbf{SSD Promotion rule.}
When an object $i$ leaves DRAM, the oracle decides whether to write it to SSD or drop it.
Because the oracle knows the future, it knows $H_i$, the number of times object $i$ will be requested again.
Let $x_i = 1$ if the oracle writes $i$ to SSD and $x_i = 0$ if it drops it.
The total cost of object $i$ is then
\begin{equation}
    C_i(x_i) = \underbrace{(1-x_i)\,H_i\,C_r}_{\text{recompute if dropped}}
             + \underbrace{x_i\,C_w\,s_i}_{\text{write if kept}} .
    \label{eq:object-cost}
\end{equation}
The cheaper option wins, so the oracle writes an object only when the recompute benefit worth more than the endurance cost.

\textbf{Eviction when the SSD is full.}
If the admission causes an eviction, there is additional replacement cost of losing the future hits of the evicted objects.
The oracle picks the residents with the fewest future hits as eviction candidates.
It writes the new object only if its benefit, $H_i C_r - C_w s_i$, is larger than the recompute cost of the hits lost by evicting the victim candidates, i.e., $H_i - \lambda s_i > H_{\mathrm{left}}$, where $H_{\mathrm{left}}$ sums the remaining hits of the evicted candidates.
Otherwise it keeps the SSD as it is and drops the new object.

\subsection{Capacity-aware KV Cache Management}
\label{subsec-capacity}


\sys keeps the oracle's objective~\eqref{eq:oracle-cost}: maximize the recomputation saved by cache hits while minimizing the cost of SSD writes and replacement cost of evicted objects.
With the CPU/SSD ratio insight in \S\ref{subsec-understanding}, we design a practical policy that approximates the oracle's decisions while considering the occupancy of each tier in an online manner.

\textbf{Promotion decision.}
Every admitted KV cache is first placed in DRAM, which costs no endurance.
When DRAM is full, \sys prefers moving the cache $j$ to SSD when dropping causes recomputation that exceeds the cost of writing it to SSD:
\begin{equation}
    \alpha \cdot move\_cost_j \;<\; drop\_cost_j
    \quad\Leftrightarrow\quad
    \frac{drop\_cost_j}{move\_cost_j} \;>\; \alpha ,
    \label{eq:gate}
\end{equation}
and otherwise drops it.
$\alpha$ is a tunable hyperparameter.
Larger $\alpha$ makes writes more selectively, reducing SSD wear but increasing cache misses, whereas smaller $\alpha$ does the opposite.

\subsubsection{Dropping a DRAM candidate}

Dropping an item misses the cache hit opportunity, so it costs the recomputation of the dropped object on its next read, which is the factor of number of hits $H_d$, the chunks have to be recomputed $v_d$, and the GPU cost to recompute one chunk $C_r$.
However, in online setting, $H_d$ is unknown, so \sys estimates it from the past history of requests.
Three observations make the estimation simple.

\begin{figure}[t]
    \centering
    \includegraphics[width=170pt]{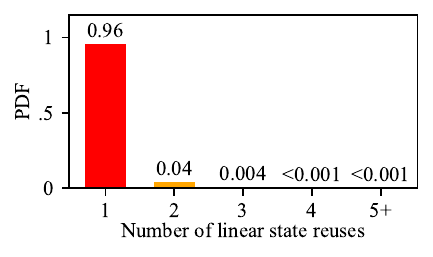}
    \caption{In Claude Code WEKA trace, most linear-attention states are reused once or never.}
    \label{fig-linear-histogram}
\end{figure}

\textit{First, only the next read matters.}
A drop decision directly loses the \emph{next} hit.
When it is reused again, the next read is recomputed and admitted back to DRAM, where \sys decides its fate again.
When \sys decides its fate, it already pays for the recomputation cost of the object that it causes to be evicted.
We therefore needs to know only \emph{whether} $d$ will be read again, not how many times.
Linear attention states is simple to predict: as shown in {\bf Figure~\ref{fig-linear-histogram}} most linear states are read once and then discarded, so $H_d$ is either one or zero.
We therefore only need to predict next reuse probability for full attention KV chunks.

\textit{Second, returns are predictable from history.}
In the history, there has been many requests being processed, and we can learn from them to predict the future.
Suppose the chunk's session has been idle for $a_d$ requests.
\sys asks: of the past sessions that were also idle this long, how many came back?
The chance of return is the share that came back, denoted as $H_d$.
It is close to 1 when the predictor, based on the past experience with many requests, sure that the cache will be read again, close to 0 otherwise.


\textit{Third, an object needed soon has a high value.}
Keeping $d$ is not free: it holds $s_d$ bytes of DRAM until its next read.
Thus, evicting an object that is needed farthest in the future is the cheapest, because it frees the most DRAM for the longest time~\cite{belady1966study}.
In the same light, an object idle for long is likely to stay idle for long.
Therefore, \sys divides~\eqref{eq:drop-cost} by $s_d$ and by $a_d$, the wait time of $d$ so far.
It prices a drop by its cost per byte of DRAM per request it frees, shown in {\bf Equation~\ref{eq:drop-cost}}.
\begin{equation}
    drop\_cost = H_d\, v_d\, C_r
    \;\to\;
    h_d = \frac{H_d\, v_d\, C_r}{s_d\, a_d}.
    \label{eq:drop-cost}
\end{equation}

\subsubsection{Moving a DRAM candidate to SSD}

Moving an object to SSD costs the SSD's endurance, plus the future hits lost on the objects evicted to make room for it.

\textit{SSD write cost.}
Writing $s_j$ bytes to SSD consumes $C_w s_j$ dollars of endurance, where $C_w = P_{\mathrm{SSD}} / E_{\mathrm{TBW}}$ is the SSD's purchase price per byte of rated write endurance.
To make SSD lasts for $T_{\mathrm{target}}$ years, \sys limits the bytes written to SSD at elapsed time $t$, so the write cost is scaled by a multiplier $t_L$ that grows when the bytes written $W$ exceeds the budget at time $t$, and shrinks when they fall behind $W$.
Without a target, $t_L = 1$ and the price alone limits wear.

\emph{Replacement cost.}
Promoting $j$ may evict other objects $E_j$ from SSD.
Each of the evicted objects might still be needed in the future, losing its remaining $H_e$ reads, each worth $v_e C_r$, as in~\eqref{eq:drop-cost}.

\textit{Total move cost.}
Following the third observation, \sys prices a move by its cost per byte of DRAM per request it frees, so that it compares with $h_d$:
\begin{equation}
    move\_cost =
    \underbrace{C_w\, s_j}_{\text{cost of writing } j}
    \,+\, \underbrace{\textstyle\sum_{e \in E_j} H_e\, v_e\, C_r}_{\text{replacement cost in SSD}},
    \,\to\,
    \frac{t_L\, C_w}{a_j}
    \,+\,
    \min_{e \in E_j} h_e
    \label{eq:move-score}
\end{equation}
\sys evicts the SSD leaves with the lowest $h_e$ and charges the cheapest of them.
While the SSD has room, $E_j$ is empty and replacement cost is not charged.

\subsection{Optimization}
\label{subsec-optimization}


Scoring every cached chunk for every eviction is expensive: the cache holds millions of chunks, and their probability-weighted scores evolve over time.
\sys avoids this cost with three observations.

\textit{First}, only leaves can be dropped, and only root-side DRAM nodes can be moved to SSD.
\sys only maintains these two sets in an incremental manner.
Dropping a leaf exposes at most its parent as a new leaf, while moving a root-side node exposes its children as new move candidates.
Thus, recomputing the scores of the caches only happens when there is cache pressure, to a small number of items.


\textit{Second}, scores in Equation~\eqref{eq:drop-cost} change as objects age, but remain fixed within a request.
\sys keeps each tier's leaf candidates in a min-heap ordered by their probability-weighted drop scores $H_d$.
A heap is rebuilt only when first queried in a new request and is reused for all subsequent evictions in that request.
New leaves are inserted immediately, while stale entries are discarded lazily from the heap top.


\textit{Third}, \sys partitions root-side DRAM candidates by cache type: linear and full.
Within each size class, the candidate with the longest estimated remaining life has the lowest SSD write fee.
A lazy heap ordered by remaining life therefore identifies this candidate without rescoring the class again during the request.
By default, \sys compares only the longest-lived candidate from each size class.
In later evaluation, we show that comparing only one candidate does not sacrifice performance.
A separate lazy min-heap over SSD leaves identifies the lowest-valued SSD victim, and hence the replacement cost when DRAM pressure triggers a move from DRAM to SSD such that an object needs to be evicted from SSD to make room.
After the heaps are built, each decision examines only their heads, rather than rescoring or sorting all cache objects.

\section{Implementation}


\textbf{Verifying the simulator.}
We first ran vLLM~\cite{kwon2023vllm} and connect it with LMCache~\cite{cheng2025lmcache} on RawBlock mode, with a Claude Code trace, then attaching an eBPF-based logger to dump out the storage-facing trace.
This includes columns such as: timestamp since first run, type of operation (R/W), logical block address (LBA) accessed, size, and number of pages.
We then replayed this trace on FEMU~\cite{li2018femu} with a 1TB @ 0.3 DWPD for 5 years SSD configuration to observe SSD lifetime given a KV cache offloading workload, resulting to our finding that KV cache is killing SSD.
From this result, we obtained the SSD lifetime duration.
Then, we build a trace-based simulator that estimates SSD lifetime by the TBW, and confirm the lifetime using the SSD lifetime we obtained by using SSD emulator.
Simulator are needed to test all of our experiments prior to implementing in real hardware, which will effectively kills SSD for sure.
After all technique was confirmed, we move our technique to real production system, involving LMCache~\cite{cheng2025lmcache} deployed as a backend for vLLM~\cite{kwon2023vllm} inference engine.


\textbf{SSD lifetime estimation.}
Specifically, given that we confirmed the SSD lifetime using an emulator in the small scale, we estimate SSD lifetime by the TBW (Total Bytes Written).
Once one trace is replayed, we can obtain the total write volume to SSD.
Assuming that this trace is replayed over and over in production system, we can then obtain the total write volume per day, and then estimate the SSD lifetime by dividing the TBW by the total write volume per day.


\textbf{Transfer between tiers.}
Transfer between one tier and another assumes modes two-hops, using CPU as a staging buffer.
Hit served from SSD must be transferred to CPU DRAM adhering SSD read bandwidth and SSD transfer bandwidth; hit served from CPU DRAM will go to GPU memory adhering CPU-to-GPU PCIe bandwidth;


\textbf{Hardware specs.}
The simulator models the hardware specs of the system that affects the transfer between SSD, CPU DRAM, and GPU HBM.
We assume a specific GPU model, check the tokens per second throughput, and every generated KV cache is directly saved in the KV cache storage, \eg LMCache~\cite{cheng2025lmcache}.
For the SSD model, the simulator supports specifying DWPD (Drive Writes Per Day) and capacity.
However, for our evaluation, we assume 1 DWPD (Drive Writes Per Day) for 5 years, which is a majority specification for enterprise SSDs.
We check the SSD read and transfer bandwidth via FIO~\cite{web.fio}.
We then vary the SSD capacity for the evaluation.

\textbf{Known limitation.}
The simulator assumes write-amplification factor (WAF) to be 1, since the simulator does not model the SSD internal operations, such as garbage collection, wear leveling, and other internal operations that may affect the SSD lifetime.
Hence, the simulator may overestimate the SSD lifetime for all the evaluated techniques, \sys and baselines.
However, since WAF is implicated by the fast write rate, if we reduce the write volume using \sys, we can still reduce the WAF and improve the SSD lifetime.




\section{Evaluation}

\subsection{Setup}



\textbf{Baselines.}
We compare \sys against six cache-management baselines: 
\textit{general tier eviction policies} such as LRU, Marconi~\cite{pan2025marconi}, Workload-aware~\cite{wang2025workloadaware};
\textit{general tier selective admission policy} such as MeritKV~\cite{khemani2026meritkv};
and \textit{GPU to DRAM selective admission policies} such as HotPrefix~\cite{li2025hotprefix} and LRU-K~\cite{peng2026lruk};


Since not all baselines are designed to operate with DRAM and SSD, for evaluation fairness, we augment each baseline with both tiers as a capacity expansion to GPU HBM, thereby increasing their effective cache capacity while preserving their original admission and victim selection rules.
A KV cache entry is admitted to lower tiers only if it is evicted from the upper tier.
Eviction happens as a side effect of admission, and the eviction policy of each baseline is retained as is.
For all baselines that has hyperparameters, we sweep the hyperparameter space to show the best configuration on each trace.

\begin{table}[t]
\centering
\small
\caption{The traces writes exceed the SSD's workload DWPD rating.}
\label{tab-trace-stats}
\begin{tabular}{llrr}
\toprule
 & & \multicolumn{2}{c}{Required DWPD} \\
\cmidrule(lr){3-4}
Trace & Model & 1TB & 4TB \\
\midrule
\texttt{cc-traces-weka-062126} & Qwen3.5-27B & 34.4 & 8.60 \\
 & Kimi-K3 & 48.2 & 12.0 \\
\addlinespace
\texttt{qwen\_traceA\_blksz\_16} & Qwen3.5-27B & 130 & 32.4 \\
 & Kimi-K3 & 274 & 68.5 \\
\addlinespace
\texttt{aiperf-100r-500o-noshare} & Qwen3.5-27B & 278 & 69.4 \\
 & Kimi-K3 & 579 & 145 \\
\bottomrule
\end{tabular}
\end{table}

\textbf{Workload Traces.}
We evaluate \sys on real-world traces with known timestamps and request sequences, including:
\begin{enumerate}
    \item \textit{Qwen Bailian Trace A}~\cite{wang2025workloadaware}, the trace from Aliyun Bailian inference service that contains 46.4\% multi-turn conversations with text, file understanding, multimodal understanding, and search tasks;
    \item \textit{Claude Code WEKA SemiAnalysis trace}~\cite{semianalysisai_cc_traces_weka_062126}, Claude Code agentic serving traces showing multi agents, subagents, and compacted traces;
    \item \textit{AIPerf Trace}~\cite{web.aiperf}, a trace generated using the AIPerf benchmark tool that contains 400k requests, where each session contains 100 turns of user queries and responses.
\end{enumerate}
The statistics of the traces are summarized in {\bf Table~\ref{tab-trace-stats}}.

\textbf{Models and Cache Sizes.}
Throughout the experiments, we evaluate \sys and the baselines on Kimi-K3, a representative hybrid model having 72\% linear attention layers, a typical ratio for many hybrid LLMs.
We set the DRAM cache size to 100GB and vary the SSD size: 1TB, 4TB, following the common SSD capacity range for enterprise SSDs.
This DRAM size is chosen to hold less than 1\% of the peak active KV cache size of most of the traces shown in Table~\ref{tab-trace-stats}, which is a reasonable assumption for a production system with limited DRAM capacity.

\begin{figure*}[t]
    \centering
    \includegraphics[width=\textwidth]{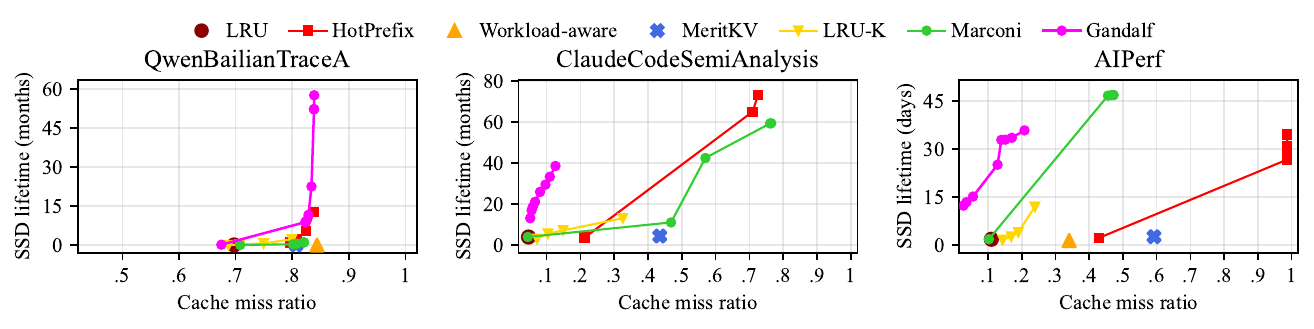}
    \caption{SSD lifetime against cache miss ratio for \sys and the baselines on all traces with Kimi-K3 model, 100\,GB of DRAM and a 1\,TB SSD. \sys, LRU-K, HotPrefix and Marconi are lines through their hyperparameters (\sys's $\alpha$, LRU-K's $K$, HotPrefix's hotness frequency, and Marconi's $\alpha$).}
    \label{fig-eval-e2e-miss}
\end{figure*}

\textbf{Metrics.}
We evaluate \sys and the baselines on the following metrics:
\begin{itemize}
    \item \textit{SSD lifetime}, measured in years, is the time until the SSD reaches its write endurance limit, given the total bytes written (TBW) and the drive writes per day (DWPD) specification.
    \item \textit{Cache miss ratio}, the fraction of requests that are served from the cache (DRAM or SSD) without needing to recompute the KV cache entries.
    \item \textit{Cache load latency}, measured in seconds, is the time taken to load KV cache from DRAM and SSD to GPU, or to recompute the KV cache entries if they are not present in the cache.
\end{itemize}


\subsection{End-to-end Performance}

{\bf Figure~\ref{fig-eval-e2e-miss}} compares \sys with the baselines on TraceA, ClaudeCode and AIPerf with the Kimi-K3 model, 100\,GB of DRAM and a 1\,TB SSD.
\sys, as well as other baselines with trade-off hyperparameters, are shown as lines through their hyperparameter space.
\sys uses the write-fee multiplier $\alpha$ (0--64), LRU-K uses $K$ (1--10), HotPrefix uses the frequency threshold $F$ (1--10), and Marconi uses its own $\alpha$ (0--2).

\textbf{SSD lifetime.}
In this hardware configuration, at its lowest miss ratio ($\alpha=0$), \sys extends the SSD lifetime over the lowest-miss baseline by 2\tms on TraceA, 3.3\tms on ClaudeCode, and 6.5\tms on AIPerf.
It does so by writing 2.0--6.5\tms fewer bytes to the SSD.
Raising $\alpha$ trades hits for a longer lifetime.
On ClaudeCode, every baseline point with a miss ratio of up to 0.47 lasts 2.9--10.3\tms shorter than \sys at an equal or lower miss ratio.
Only Marconi with $\alpha\ge1$ and HotPrefix with $F\ge7$ last longer than \sys's longest-lived point, and they miss 4.5--6.1\tms more.
On AIPerf, every baseline point except Marconi with $\alpha\ge0.5$ lasts 1.0--23.8\tms shorter than \sys at an equal or lower miss ratio.
Marconi last 1.3\tms longer than \sys's longest-lived point, but results in 2.2\tms more cache misses.
On TraceA, \sys reaches 57.7 months at a 0.839 miss ratio, 4.1\tms the 12.7 months of the longest-lived baseline, HotPrefix with $F=10$, at the same miss ratio.
LRU-K works in the opposite direction, where increasing $K$ trades a lower miss ratio for a longer lifetime.

\textbf{Cache miss ratio.}
\sys reaches the lowest miss ratio of all policies on TraceA and on AIPerf.
On ClaudeCode, Marconi with $\alpha=0$ and LRU miss slightly less than \sys (0.045 and 0.047 vs.\ 0.052), but write 3.3\tms more to the SSD.

\textit{In summary, with DRAM size of 100GB and 1TB SSD, \sys~improves SSD lifetime by 7.3\tms on average while maintaining or even reducing the cache miss ratio.}


\subsection{Cache Ratio Sensitivity}


\begin{figure}[t]
    \centering
    \includegraphics[width=190pt]{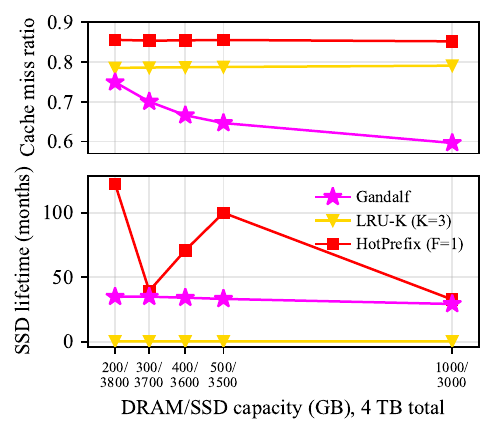}
    \caption{\sys achieves lower cache miss ratio while achieving higher SSD lifetime under different DRAM/SSD capacity ratios. We fix the total capacity of DRAM and SSD to 4TB to make sure the cache miss ratio and lifetime is comparable across different setups.}
    \label{fig-eval-capacity}
\end{figure}

{\bf Figure~\ref{fig-eval-capacity}} shows TraceA with Kimi-K3 under different DRAM/SSD capacity ratios, keeping the total DRAM+SSD capacity at 4TB, from 200GB of DRAM with a 3.8TB SSD to 1000GB with a 3TB SSD.
\sys is ran while keeping write price per GB constant, so the decision to write to SSD is only affected by the capacity ratio.
\sys uses the extra DRAM to keep more KV cache entries in memory, which reduces the miss ratio from 0.75 to 0.6, while its SSD lifetime stays similar across all ratios.
Its SSD lifetime stays between 29 and 35 months across all ratios, decreasing slightly only because the SSD size itself shrinks.
The baselines barely benefit from the extra DRAM. LRU-K's miss ratio stays at 0.785--0.791.
HotPrefix, while it increases SSD lifetime to 33-122 months, it has an impractical miss ratio of 0.852--0.856, which does not meet deployment Service Level Objectives (SLOs) because it is too far from the trace's maximum achievable miss ratio of 0.6.
LRU-K writes a KV entry evicted from DRAM to the SSD when the session's $K$-th most recent entry is still in DRAM, hence relying on the past historical access without considering whether the entry will be reused in the future.
On TraceA most entries pass it, so LRU-K writes about 60TB to the SSD and keeps both tiers full; with the total capacity fixed, moving capacity from the SSD to DRAM changes little, and its miss ratio stays at 0.785--0.791.
\sys instead writes only about 0.5TB to the SSD, keeping most of its working set in DRAM where writes cost no SSD wear, so additional DRAM turns directly into hits.

Hence, \textit{\sys improves cache miss ratio by 5--25\% over the best baseline at every capacity ratio, while extending SSD lifetime by 120\tms in various tested scenarios with total cache capacity at 4TB.}


\subsection{Trade-Off Hyperparameter Sensitivity}\label{subsec-eval-hyperparam}

{\bf Figure~\ref{fig-eval-e2e-miss}} also sweeps $\alpha$, the weight \sys gives to the move cost when deciding whether to write an entry to the SSD~\eqref{eq:gate}, from 0 to 64, with 100GB of DRAM and a 1TB SSD.
By default, \sys sets $\alpha = 1$, which balances SSD lifetime and cache hit ratio across all traces.
A larger $\alpha$ makes writing to the SSD more selective, so the SSD lives longer at the cost of more misses: from $\alpha = 1$ to $\alpha = 64$, the SSD lifetime grows 6.5\tms on TraceA, 2.3\tms on ClaudeCode, and 2.7\tms on AIPerf (0.4 to 1.2 months), while the miss ratio rises by 0.016, 0.071, and 0.172, respectively.
A lower $\alpha$ lets \sys trade lifetime for hits.
With $\alpha = 0$, \sys wins in both metrics against the baseline with the lowest miss ratio on TraceA and on AIPerf.
On ClaudeCode, \sys at $\alpha = 0$ comes within 0.007 of Marconi's miss ratio while its SSD lasts 3.3\tms longer.

\textit{Therefore, operators can tune $\alpha$ to meet their desired trade-off between SSD lifetime and cache hit ratio, and \sys is robust to a wide range of $\alpha$ values.}

\subsection{Model Linear-to-Full Sensitivity}

\begin{figure}[t]
    \centering
    \hspace*{-10pt}\includegraphics[width=250pt]{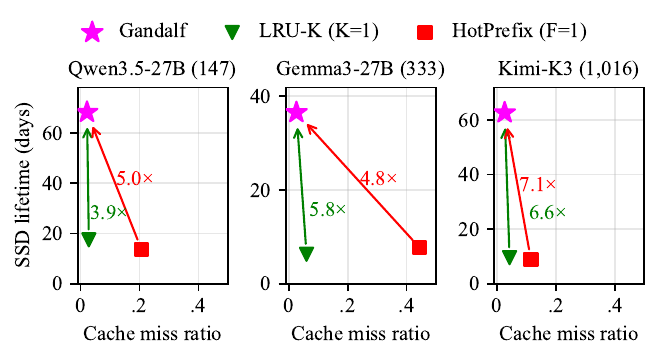}
    \caption{
    \sys improves baselines in wide range of hybrid models having different linear-to-full cache size ratios.
    }
    \label{fig-eval-model}
\end{figure}



{\bf Figure~\ref{fig-eval-model}} compares \sys with LRU-K and HotPrefix on AIPerf for four hybrid models with 100GB of DRAM and a 4TB SSD, ordered by the size of a linear-attention state relative to a full-attention KV chunk, from 147\tms for Qwen3.5-27B to 1{,}016\tms for Kimi-K3.
\sys extends the SSD lifetime on every model, by 3.6--6.6\tms over LRU-K and 3.6--7.1\tms over HotPrefix.
It also reduces the miss ratio on three of the four models, by 0.006--0.034 relative to LRU-K and by 0.089--0.419 relative to HotPrefix.
On Nemotron-H, \sys keeps the miss ratio similar to LRU-K, but extends the SSD lifetime by 3.6\tms.
Among Nemotron-H, Gemma3-27B, and Kimi-K3, the lifetime gain over LRU-K grows with the state-to-KV ratio, from 3.6\tms to 5.8\tms and 6.6\tms, and Qwen3.5-27B.
Even in the model with the smallest ratio, \sys still gains 3.9\tms.

\textit{The larger the ratio, the higher the penalty of writing a linear state that is never reused, so the improvement obtained by \sys's filtering is larger.}
This makes \sys especially needed for deployment of hybrid models with large linear-attention states, which are increasingly common in modern LLMs.





\subsection{Where does the improvement come from?}

\begin{figure}[t]
    \centering
    \includegraphics[width=200pt]{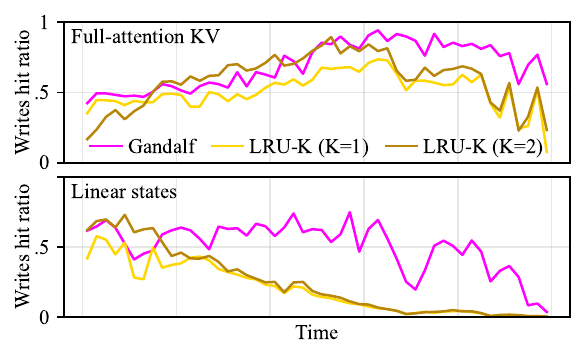}
    \caption{\sys SSD admission results in higher fraction of SSD writes that contributes to cache hits at least once. Unlike baselines, \sys cache hit is stable across time.}
    \label{fig-eval-ablation}
\end{figure}


\textbf{Avoid writing non-reused linear states in SSD.}
\sys avoids writing linear states that will not be reused multiple times, such as those at every turn of a multi-turn conversation, to the SSD.
{\bf Figure~\ref{fig-eval-ablation}} shows, over time, the fraction of SSD writes that contribute to cache hits at least once, for \sys and LRU-K on ClaudeCode with Qwen3.5-27B, 100GB of DRAM and a 1TB SSD.
Throughout the trace, a higher fraction of \sys's SSD writes are hit than those of LRU-K at $K=1$, for both full-attention KV and linear states.

LRU-K at $K=1$ writes 120.6k linear states to the SSD, and only 21\% of them are ever hit there: 43\% are never requested again, and 45\% had already served their one hit from DRAM before being moved to the SSD, i.e., the state of a turn that the next turn already reused.
A larger $K$ does not fix this.
LRU-K~\cite{peng2026lruk} counts references per object, which captures hotness for objects that persist across requests, such as a user's KV cache in generative recommendation.
A linear state, however, is superseded at every turn and is referenced at most about twice, so it reaches $K=2$ references only after its one reuse is over: LRU-K drops a state that leaves DRAM before the next turn arrives, and writes one that the next turn has already reused.
With $K=2$, LRU-K writes 1.9\tms fewer linear states (62.6k), but 90\% of them had already been hit before being written, 58\% are never requested again, and only 13\% are hit on the SSD; meanwhile its miss ratio rises from 0.1 to 0.14.
\sys instead decides by whether a state will be reused later, not by how often it was referenced: it writes 8.8\tms fewer linear states than LRU-K at $K=1$, and 57\% of them are hit on the SSD.

As the SSD is full, LRU-K gets more selective at admitting cache to SSD, hence the cache hit rate decreases.
\sys, however, maintains a high cache hit rate throughout.

\textbf{Avoid writing rarely reused full-attention KV in SSD.}
We also increase the reusability of the full-attention KV cache entries that are written to the SSD, by filtering out those that are unlikely to be reused in the future, such as those at the end of a conversation or session.
\sys writes 33\% fewer full-attention KV chunks than LRU-K at $K=1$ (21.5M vs.\ 32.1M), and 61\% of them are hit on the SSD, compared to 50\% for LRU-K.
Of the chunks LRU-K writes, 9.6\% are never requested again, compared to 5.3\% for \sys, so \sys writes 2.7\tms fewer such chunks (1.1M vs.\ 3.1M).
Despite writing fewer chunks, \sys serves 18\% more full-attention hits from the SSD (265M vs.\ 225M), i.e., 12.3 hits per chunk written compared to 7.0 for LRU-K.
Together, \sys writes 92TB to the SSD instead of 153TB, which extends the SSD lifetime by 1.7\tms (98 vs.\ 59 days) at a similar miss ratio (0.099 vs.\ 0.098).



\begin{figure}[t]
    \centering
    \includegraphics[width=140pt]{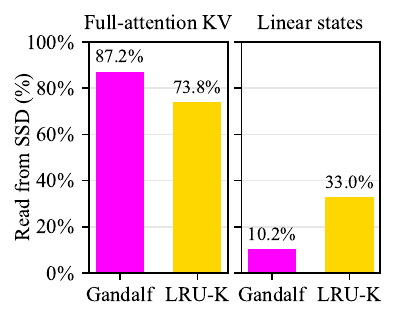}
    \caption{
    A bigger portion of linear state cache hit in \sys comes from DRAM, while most full attention KV cache hit comes from SSD.
    }
    \label{fig-eval-occupancy}
\end{figure}

\textbf{Attention type-aware placement.}
\sys places linear states in DRAM and full-attention KV in the SSD.
{\bf Figure~\ref{fig-eval-occupancy}} shows where the cache hits of each kind are served from, for \sys and LRU-K at $K=1$ in the same setting as above.
\sys serves 87\% of its full-attention KV hits from the SSD and 90\% of its linear-state hits from DRAM.
LRU-K, which places objects by recency regardless of their kind, serves 74\% of its full-attention KV hits from the SSD and only 67\% of its linear-state hits from DRAM since its DRAM also holds full-attention KV that \sys would have moved to the SSD.
To serve the remaining 33\% (27.0k hits) from the SSD, LRU-K writes 120.6k linear states to it, 4.5 writes per hit, whereas \sys serves 8.0k linear-state hits from the SSD with 13.7k writes, 1.7 writes per hit.

\textit{In summary, the improvement of \sys comes from leveraging attention type as a proxy for reusability, which: (1) avoids writing linear states at the end of each rounds to the SSD, (2) avoids writing rarely reused full-attention KV to the SSD, and (3) places linear states in DRAM and full-attention KV in the SSD.}

\begin{figure}[t]
    \centering
    \includegraphics[width=\columnwidth]{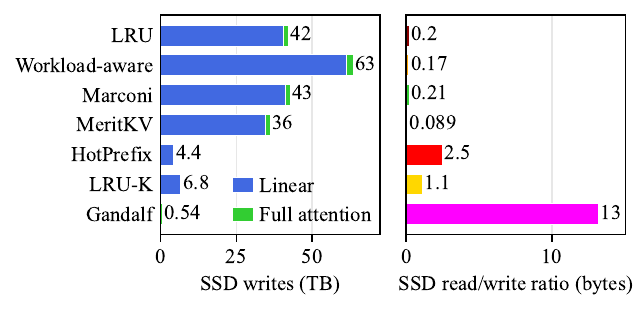}
    \caption{The gain from \sys comes from both reducing the amount of writes to the SSD (left) and increasing the Read/Write ratio (right).}
    \label{fig-eval-breakdown}
\end{figure}

\subsection{Comparing \sys to each baseline.}
{\bf Figure~\ref{fig-eval-breakdown}} breaks down each policy's SSD writes on TraceA (2 hours duration) with Kimi-K3, 100GB of DRAM and a 1TB SSD, with \sys at $\alpha=1$, HotPrefix at $F=1$, LRU-K at $K=3$, and Marconi at $\alpha=0$.

\textit{Write-back baselines (LRU, Workload-aware, Marconi).}
These policies write every object evicted from DRAM to the SSD: they decide which object to evict, but not whether to move it to the SSD or drop it.
Linear states make up 96\% of the bytes they write, yet only 1.6--8.5\% of the linear states they write are ever hit on the SSD, and 57\% of LRU's are never requested again.
\sys writes 25 linear states instead of LRU's 90k, and 3.2\tms fewer full-attention bytes (0.53 against 1.70TB).
In total, it writes 73\tms fewer bytes than LRU (0.58 against 42.2TB), and its SSD lasts 73\tms longer (253 against 3.5 days).
Workload-aware and Marconi change only which object to evict; Marconi writes about as much as LRU (42.7TB, 3.4 days), and Workload-aware writes 1.5\tms more (63.5TB, 2.3 days).
At $\alpha=1$, \sys has a higher miss ratio than LRU (0.818 against 0.697; see below).
With $\alpha=0$, \sys has the lowest miss ratio of all policies, 0.603, and its SSD still lasts 1.4\tms longer than LRU's (4.95 against 3.46 days), because what it writes is read more: its SSD serves 1.8\tms as many full-attention hits (2.04M against 1.14M) and 2.2\tms as many linear-state hits (38.9k against 18.0k).

\textit{MeritKV.}
MeritKV admits an object to the SSD only when the prefill it saves outweighs the expected reuse latency plus the write cost, but it admits the first three demands of every object unconditionally to learn its reuse.
Most linear states are written before this estimate applies: 97\% of the linear states it writes had never been hit before the write.
It still writes 36.1TB (4.0 days), only 5.6\% of its linear-state writes are hit on the SSD, and it misses 0.807 of chunks.

\textit{HotPrefix.}
HotPrefix admits a prefix to DRAM only if it has been requested at least $F$ times while DRAM has room, and otherwise only if its frequency times its recency clock beats the coldest DRAM entry; a refused prefix is dropped rather than written to the SSD.
This cuts its writes to 4.4TB, 9.6\tms fewer than LRU, and its SSD lasts 33 days.
The filter, however, does not tell reused linear states from the rest: 97\% of the linear states it writes had never been hit before, 51\% are never requested again, and only 17\% are hit on the SSD.
The dropped prefixes become misses, so its miss ratio is 0.802; \sys at $\alpha=1$ lasts 7.6\tms longer at a 0.016 higher miss ratio.
With $F=10$, HotPrefix writes 0.38TB and lasts 388 days, but misses 0.839 of chunks.

\textit{LRU-K.}
LRU-K writes an object to the SSD only once it has been referenced $K$ times.
With $K=3$, it writes 6.8TB, 6.2\tms fewer than LRU, and 31\% of the linear states it writes are hit on the SSD, against 7.9\% for LRU.
The objects it refuses are dropped, so its miss ratio rises to 0.750, and its SSD lasts 21.5 days; \sys at $\alpha=1$ lasts 11.8\tms longer at a 0.068 higher miss ratio.


\textit{Where \sys gives up hits.}
At $\alpha=1$, \sys misses more than LRU on TraceA (0.818 against 0.697).
Since it writes only 25 linear states, a conversation whose linear state has left DRAM must be recomputed from the start, and its full-attention KV cannot be used either: \sys serves 19.5k linear-state hits against 42.4k for LRU, and 1.11M full-attention hits against 1.91M.
In exchange, its SSD lasts 73\tms longer.
A smaller $\alpha$ recovers these hits at a shorter lifetime (\S\ref{subsec-eval-hyperparam}).

\subsection{How much is the cache load latency affected?}

\begin{figure}[t]
    \centering
    \includegraphics[width=230pt]{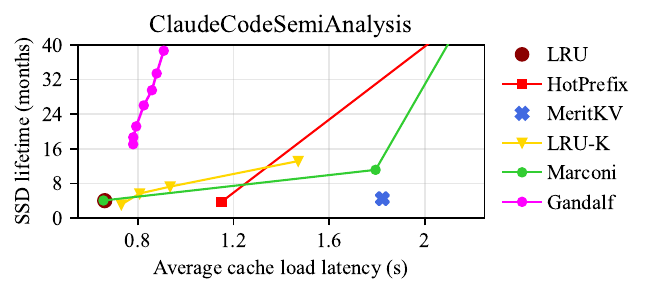}
    \caption{\sys increases SSD lifetime while maintaining low cache load latency.}
    \label{fig-eval-ablation-ttft}
\end{figure}

We approximate the cache load latency of a request as the maximum of the time to transfer its cached KV to GPU HBM and the time to recompute its missed tokens, since the two overlap.
The transfer uses DRAM as a staging buffer: KV on the SSD is first read into DRAM (10GB/s), then every cached byte is sent to GPU HBM (50GB/s); recomputation runs at 60k tokens/s.
{\bf Figure~\ref{fig-eval-ablation-ttft}} shows the average cache load latency against SSD lifetime on ClaudeCode with Kimi-K3, 100GB of DRAM and a 1TB SSD.
The baselines with the lowest load latency, Marconi at $\alpha=0$ and LRU, load in 0.66s and their SSDs last 4 months.
\sys at $\alpha=1$ lasts 4.2\tms longer (17.0 months) at a load latency of 0.78s, 0.12s higher.
It reads more of its hits from the SSD, which is slower than DRAM, and its miss ratio is 0.01 higher (0.056 against 0.045), so slightly more tokens are recomputed.
At a similar load latency, \sys lasts 3.7\tms longer than LRU-K at $K=2$ (21.2 against 5.7 months, 0.79s against 0.81s), and with $\alpha=64$ it lasts 5.3\tms longer than LRU-K at $K=3$ while still loading faster (38.6 against 7.3 months, 0.91s against 0.93s).

\begin{figure}[t]
    \centering
    \includegraphics[width=200pt]{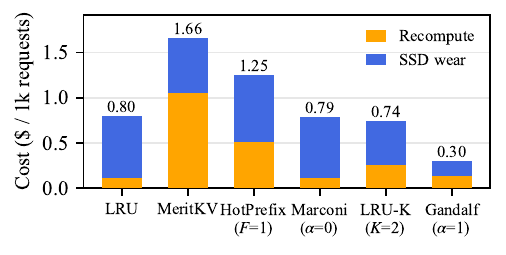}
    \caption{Average cost per request, recomputation plus SSD wear, on ClaudeCode with Kimi-K3, 100\,GB of DRAM and a 1\,TB SSD. \sys has the lowest cost once the SSD wear is counted.}
    \label{fig-ablation-price}
\end{figure}

However, more recomputation is not always worse.
Every write wears out the SSD, effectively using a portion of its purchase price, so the cost of a request is the GPU cost of recomputing its missed tokens plus the SSD price share.
{\bf Figure~\ref{fig-ablation-price}} shows this cost as an average per request, choosing the cheapest hyperparameters for each policy.
SSD wear makes up 66--86\% of the cost of LRU, Marconi and LRU-K, and the cheapest baseline, LRU-K at $K=2$, costs \$0.74 per 1k requests.
\sys at $\alpha=1$ costs \$0.30 per 1k requests, 2.5\tms less: it recomputes more than LRU and Marconi (\$0.13 against \$0.11 per 1k requests), but its SSD wear is 3.0\tms lower than that of any baseline (\$0.16 against \$0.49--0.73).

\textit{Therefore, \sys may choose to recompute more tokens if the SSD write cost is high, since the inference saving cost involves not only GPU as a hardware resource, but also SSD.}

\subsection{The optimization does not affect the performance negatively.}

To keep each decision cheap, the move door~(\S\ref{subsec-optimization}) checks, for each slot size, only the longest-lived DRAM entry as a candidate to move to the SSD.
A move is then blocked whenever that one entry's score falls below its write fee, even if another entry of the same size would be worth writing.
To measure what this costs, we let the move door check up to 64 candidates per slot size, with Kimi-K3, 100GB of DRAM, a 1TB SSD and $\alpha=1$.
Checking more candidates does not lower \sys's miss ratio: it stays at 0.823 on TraceA and 0.036 on AIPerf for every number of candidates.
On TraceA, the extra candidates only move entries that are not read again, so the SSD writes grow by 1.4\% (501 to 508GiB) and the SSD lifetime shortens from 271 to 268 days; on AIPerf, the first candidate is never blocked, so the result is identical.
A single candidate per slot size therefore loses no hits.

\section{Related Work}

\textit{Tiered KV-cache systems.}
CachedAttention, FlashGen, LMCache, and Mooncake extend GPU memory with host DRAM, SSD, or distributed storage~\cite{gao2024cachedattention,jeong2025flashgen,cheng2025lmcache,qin2025mooncake}, while TENURE and placement studies decide where KV caches should reside across these tiers~\cite{noohi2026tenure,tumkur2026whereshouldkvcachelive}.
InfiniGen and Tutti instead reduce lower-tier access overhead through selective fetching and GPU-directed SSD I/O~\cite{lee2024infinigen,qiu2026tutti}.
These systems primarily optimize capacity, latency, or throughput rather than SSD endurance.

\textit{KV admission, eviction, and compression.}
Prior policies predict reuse from prefix structure, recency, frequency, semantics, queued demand, or workload history~\cite{pan2025marconi,li2025hotprefix,wang2025workloadaware,yang2026lpc,fang2026saecache,pan2026rr,xie2026peek,wang2026prefixshield}.
MeritKV, Bidaw, LRU-K, and NVIDIA KVBM further incorporate reuse utility, storage cost, or write awareness~\cite{khemani2026meritkv,hu2026bidaw,peng2026lruk,web.nvidiaKVBM}.
Orthogonal techniques reduce the footprint through token eviction, quantization, or model-level KV compression~\cite{zhang2023h2o,liu2024kivi,deepseekai2026deepseekv41flashpushinglimitskv}.
Most of these techniques do not jointly manage DRAM placement, SSD admission, and SSD eviction using attention-type reuse heterogeneity and an explicit endurance cost.

\textit{Flash-aware caching and storage.}
General flash caches control wear through admission filtering and write-rate constraints~\cite{eisenman2019flashield,yang2022cachesack,wong2024baleen}, while offline work extends B\'elady-style analysis to account for flash erasures~\cite{cheng2016erasing}.
Device- and interface-level techniques reduce physical writes through retention management, stream placement, and cache-aware reclamation~\cite{liu2012retention,luo2015warm,rho2018fstream,mcallister2024fairywren}.
For LLM serving, HiFC, FlashAccel, and HBFlex optimize flash data paths, writeback, and reclamation~\cite{jeong2026hifc,wang2026flashaccel,zhong2026hbflex}.
These mechanisms are complementary to \sys because they do not use attention semantics to value individual KV states.

\textit{Lifetime-aware KV placement.}
Most closely related, Lachesis places agent KV segments across HBM and high-bandwidth flash using lifetimes inferred from the agent harness~\cite{yang2026lachesis}.
In contrast, \sys targets host DRAM and SSD, derives reuse differences from hybrid-attention semantics without harness-specific information, and jointly considers capacity, recomputation benefit, and SSD write cost.

\section{Conclusion}

This paper presents \sys, an SSD-based KV cache offloading system that uses the attention type associated with each KV cache object to predict its likelihood of future reuse.
\sys retains short-lived linear-attention states in DRAM, while offloading full-attention KV caches with high reuse potential to SSD. 
Its placement policy balances the expected cache hits from each SSD write against the resulting device wear.
By filtering writes based on both reuse potential and wear cost, \sys preserves cache hit rates while extending SSD lifetime.
Across a range of hybrid models and hardware capacities, \sys improves SSD lifetime by 2--7.3\tms over the best baseline.
As KV caches continue to outgrow available memory, we hope \sys motivates future KV cache systems to consider not only the performance benefits of SSD offloading, but also the wear caused by each write.
We leave the extension of \sys to other types of flash-based storage as a future work.


\bibliographystyle{ACM-Reference-Format}
\bibliography{references}

\appendix

\end{document}